\documentclass[11pt,letterpaper]{article}
\usepackage{acl2015}
\usepackage[utf8]{inputenc}
\usepackage{times}
\usepackage{latexsym,graphicx}
\usepackage{url}
\newcommand{\wc}[1]{\textbf{}}
\newcommand{\bc}[1]{\textbf{}}

\newcommand{\iflong}[1]{}
\newcommand{\ifshort}[1]{}
\newcommand{\iflongelse}[2]{}
\newcommand{\uncite}[1]{}

\newcommand{\edge}[2]{{u\rightarrow{}v}}

\newcommand{\cd}[1]{\texttt{#1}}

\usepackage{subcaption}
\usepackage{balance}
\usepackage{algorithm}
\usepackage{algorithmic}

\usepackage{multirow}

\title{Bootstrapping Distantly Supervised IE using Joint Learning and Small Well-structured Corpora}
\begin{document}
\author{Lidong Bing \ \ Bhuwan Dhingra \ \ Kathryn Mazaitis \ \ Jong Hyuk Park \ \ William W. Cohen \\
School of Computer Science \\
Carnegie Mellon University, Pittsburgh, PA 15213 \\
\{lbing, bdhingra, krivard, jp1, wcohen\}@cs.cmu.edu}
%\author{Lidong Bing$^{\S}$ \ \ Mingyang Ling$^{\S}$ \ \ Richard C. Wang$^{\natural}$ \ \ William W. Cohen$^{\S}$ \\
%$^{\S}$Carnegie Mellon University, Pittsburgh, PA 15213\\
%$^{\natural}$US Development Center,
%Baidu USA, Sunnyvale, CA 94089\\
%{ $^{\S}$\{lbing@cs, mingyanl@andrew, wcohen@cs\}.cmu.edu} \\ {$^{\natural}$richardwang@baidu.com}}
\maketitle

\begin{abstract}
We propose a framework to improve performance of distantly-supervised relation extraction, by jointly learning to solve two related tasks: concept-instance extraction and relation extraction.  We combine this with a novel use of document structure: in some small, well-structured corpora, sections can be identified that correspond to relation arguments, and distantly-labeled examples from such sections tend to have good precision.  Using these as seeds we extract additional relation examples by applying label propagation on a graph composed of noisy examples extracted from a large unstructured testing corpus. Combined with the soft constraint that concept examples should have the same type as the second argument of the relation, we get significant improvements over several state-of-the-art approaches to distantly-supervised relation extraction.

\end{abstract}

\vspace{-3mm}
\section{Introduction}
\vspace{-2mm}

In distantly-supervised information extraction (IE), a knowledge base
(KB) of relation or concept instances is used to train an IE system.
For example, a set of facts like
\cd{side\-Effect(melox\-icam, stom\-ach\-Bleed\-ing)},
\cd{int\-eracts\-With(melox\-icam, ibu\-prof\-en)}, etc are matched against a corpus, and the
matching sentences are then used to generate training data consisting
of labeled relation mentions.  Distant supervision is less expensive to obtain than directly supervised labels, but produces noisy training data whenever matching errors occur.   Hence distant
supervision is often coupled with learning methods that allow for this
sort of noise, e.g., by introducing latent variables
for each entity mention \cite{hoffmann2011knowledge,riedel2010modeling,Surdeanu:2012:MML:2390948.2391003};
by carefully selecting the entity mentions
from contexts likely to include specific KB facts \cite{wu2010open};
   or by careful filtering of
the KB strings used as seeds \cite{Movshovitz-Attias:2012:BBO:2391123.2391126}.

Another recently-introduced approach to reducing the noise in distant supervision is to combine distant labeling with \emph{label propagation} (LP) \cite{bing-EtAl:2015:EMNLP,bing-aaai:2016}.  Label propagation is a family of graph-based semi-supervised learning (SSL) methods in which instances that are ``nearby'' in the graph are encouraged to have similar labels.  Depending on the LP method used, agreement with seed labels can be imposed as a hard constraint \cite{ZhuICML2003} or a soft constraint \cite{DBLP:conf/asunam/LinC10,DBLP:conf/aistats/TalukdarC14}. When seed-label agreement is a soft constraint, then LP can be viewed as a way of smoothing the seed labels, so that labels for groups of ``similar'' instances  (i.e., instances nearby in the graph) are upweighted if they agree, and downweighted if they disagree.

In combining distant supervision with LP, one must build a graph that connects instances that are likely to have the same label. Previously, systems have constructed graphs which connect mentions appearing in the same coordinate-list structure---e.g., the underlined noun phrases in ``Get medical help if you experience \underline{chest pain}, \underline{weakness}, or \underline{shortness of breath}'' \cite{bing-EtAl:2015:EMNLP}. This approach was shown to improve performance in recognizing instances of certain medical noun-phrase (NP) categories, such as drug names and disease names.  An extension of this approach \cite{bing-aaai:2016} learned to classify NP pairs as relations, using a more complex graph structure.

This paper presents three new contributions extending this line of work.  First, we combine the concept-instance extraction and relation-extraction tasks, in the process greatly simplifying the relation-extraction LP step.  The combination of the tasks is simple but effective.  In \cite{bing-aaai:2016}, relation extraction was performed on an ``entity centric'' corpus, where each document is primarily concerned with a particular ``title entity'', and the first argument of each relation is always the title entity: hence relation extraction can be viewed as classification, where an entity mention is labeled with its slot filling role, i.e., its relation to the title entity. The intuition behind combining concept extraction and relation extraction is that relation arguments are often constrained to be of a particular type; for example, the \cd{sideEffect} of a drug is necessarily of the type \cd{symptom}.

\begin{figure}[t]
\centerline{\includegraphics[width=0.42\textwidth]{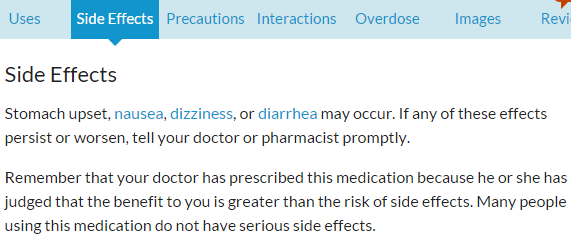}}
%\vspace{-0.1in}
\caption{\label{fig:side_effect_meloxicam}A structured document in WebMD describing the drug meloxicam.  All documents in this corpora have the same seven sections.}
\vspace{-6mm}
\end{figure}

The second contribution is a novel use of document structure; in particular, we exploit the fact that in some small, well-structured corpora, sections can be identified that correspond fairly accurately to relation arguments.
Figure~\ref{fig:side_effect_meloxicam} shows a document from such a structured corpus (discussed below) which contains sections labeled ``Side Effects''. If ``nausea'' is distantly labeled as a \cd{side\-Effect} of \cd{meloxicam} in this well-structured document, it is very likely to be a correct mention for the \cd{side\-Effect} relation. Used naively, extending a corpus with a small well-structured one needs not to lead to improvements, but when combined with LP, we show a consistent and sometimes substantial improvement in performance.  We thus illustrate a novel and effective way to make use of a small well-structured corpus, a commonly available resource that is intermediate in structure between a KB and an ordinary text corpus.

The third contribution is experimental.  We perform extensive experiments comparing this approach to state-of-the-art distant labeling methods based on latent variables, and show substantial improvements in two domains: the relative improvements under F1 measure
are from 72\% to 110\% on one domain, and 22\% to 30\% on a second domain.

Below we present our method, in outline and then in detail; present experimental results; discuss related work; and finally conclude.

\begin{figure}[t]
\centerline{\includegraphics[width=0.35\textwidth]{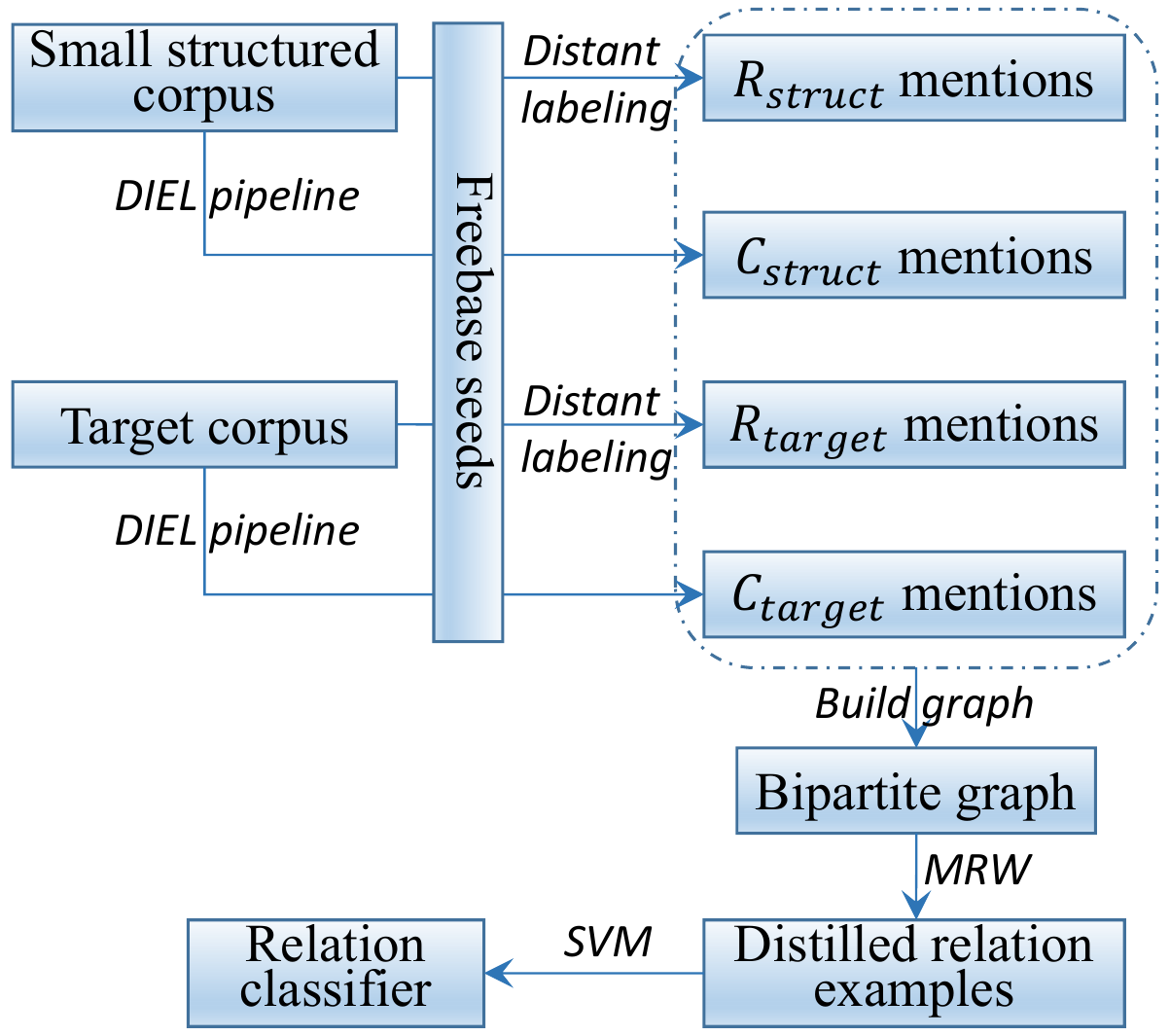}}
\caption{\label{fig:arch}Architecture of DIEJOB.}
\vspace{-5mm}
\end{figure}

\section{\mbox{
DIEJOB: \textit{\textbf{D}}istant \textit{\textbf{IE}} by \textit{\textbf{JO}}int \textit{\textbf{B}}ootstrapping}}

\subsection{Overview}
DIEJOB, our system for distantly-supervised relation extraction, is shown in Figure~\ref{fig:arch}. We consider a common case, in which most information is found in relatively unstructured free text, but some smaller corpora exist that are well-structured.
DIEJOB thus assumes at least two corpora exist for the domain of interest: a large \textit{target corpus} and a smaller \textit{structured corpus}. Further, it assumes that every document in these two corpora is associated with a particular entity, called \emph{title entity} or \emph{subject entity}. Many widely-used corpora have this structure, including Wikipedia and the authoritative consumer-oriented websites we use,  DailyMed and WebMD.

From each corpus, DIEJOB produces two types of mention sets: relation mention set $R$ and concept mention set $C$.
For the example of Figure~\ref{fig:side_effect_meloxicam}, $R$ contains a \cd{sideEffect} relation mention for ``stomach upset'' from the first sentence, and $C$ may contain mentions of the \cd{Symptom} concept, like ``stomach upset'' and ``nausea'' from the same sentence.
The tail argument values (such as ``nausea'' in \cd{sideEffect(meloxicam, nausea)}) of a relation are often from a particular unary concept. This is especially true in the biomedical domain, where for example, side\-Effect takes instances of Symptom as the value range of its second argument. Naively, those concept mentions in $C$ could serve as a source to generate relation examples,
but not all concept mentions are relation mentions: e.g., the \cd{Symptom} mentions of ``confusion'' and ``mood changes'' from ``Symptoms of overdose may include: confusion, mood changes ...'' are not mentions of the \cd{side\-Effect} relation (or any other relation we currently extract).
For the structured corpus, the relation and concept mention sets are referred to as $R_s$ and $C_s$, and for the target corpus as $R_t$ and $C_t$. Some special treatments (discussed in Section \ref{sec:example}) are done while preparing $R_s$ and $C_s$.

%mention DIEL too much seems not wise.
%In earlier work \cite{bing-EtAl:2015:EMNLP}, a system called DIEL used LP to both expand and clean a set of distantly-labeled concept instances and seeds.  We also use DIEL for postprocess the concept examples (and also generate candidate mentions).  Additionally, we use the section information to filter the examples in $C_s$ and $R_s$ fro the well-structured corpus.

After producing $R_s$, $R_t$, $C_s$ and $C_t$, DIEJOB builds a bipartite graph, following prior work \cite{lin2012scalable}, in which the nodes are either mentions in the four sets, or features of these mentions, with edges between a mention and its features.
To distill a cleaner set of relation training examples, DIEJOB performs LP on the bipartite graph.  Only the mentions from $R_s$ are used as seed relation examples in this LP stage (because they are more accurate, see Section \ref{sec:example}).

Finally the distilled relation examples are used to train an ordinary SVM classifier over their extracted features. DIEJOB thus finally learns to classify an unseen mention by the relation which holds between the mention and its corresponding title entity based on features of the mention---a convenient architecture to use for large-scale extraction.

Below we will describe the components of DIEJOB and the experiments in more detail.

\iflong{
Extensive experiments are conducted on two corpora, for diseases and drugs, and the results show that this approach significantly improves over classical distant-supervision approaches and state-of-the-art works.
}

\subsection{Relations and Corpora}
\label{sec:kb_corpus}

\iflong{Distantly-supervised IE is often used to extend an incomplete KB.  }

Even large curated KBs are often incomplete and the situation is worse in the medical domain where the coverage of large KBs like Freebase is fairly limited.
\iflong{e.g., a Freebase snapshot from April 2014 has (after filtering noise with simple rules such as length greater than 60 characters and containing comma) only 4,605 instances in the ``Disease or Medical Condition`` type and 4,383 instances in ``Drug'' type, whereas \url{dailymed.nlm.nih.gov} contains data on over 74k drugs, and \url{malacards.org} lists nearly 10k diseases.}
We focus on extracting instances of eight relations, defined in Free\-base, about drugs and diseases.
The  drug relations are \cd{used\-To\-Treat}, \cd{con\-di\-tions\-This\-May\-Pre\-vent}, and \cd{side\-Effect}, and the concept types of their second arguments are \cd{Dis\-ease\-Or\-Medi\-cal\-Con\-di\-tion}, \cd{Dis\-ease\-Or\-Medi\-cal\-Con\-di\-tion}, and \cd{Symptom}.
The  disease relations are \cd{has\-Treatment}, \cd{has\-Symptom}, \cd{risk\-Factor}, \cd{has\-Cause}, and \cd{pre\-ven\-tion\-Factor}, with  corresponding concept types as \cd{Medical\-Treat\-ment}, \cd{Symptom}, \cd{Risk\-Factor}, \cd{Disease\-Cause}, and \cd{Con\-di\-tion\-Pre\-ven\-tion\-Factor}.

We are primarily concerned with extraction from large, authoritative sources. Our target drug corpus, called DailyMed, is downloaded from {dailymed.nlm.nih.gov} and contains 28,590 XML documents. Our target disease corpus, called WikiDisease, is extracted from a Wikipedia dump of May 2015 and contains 8,596 disease articles.
The structured drug corpus\footnote{It is not difficult to
find such structured pages in different domains, such as scientist (http://famouschemists.org/, having “Famous For”, “Awards”, and “Discoveries” sections) and movie (http://www.imdb.com/chart/top, having “Awards”, “Plot Summary”, etc.)}, called WebMD, contains 2,096 pages collected from {www.webmd.com}. Each page has the same sections, such as \textit{Uses} and \textit{Side Effects}, corresponding to \cd{used\-To\-Treat}/\cd{con\-di\-tions\-This\-May\-Pre\-vent} and\cd{side\-Effect} relations, respectively.
The structured disease corpus, called MayoClinic, contains 1,117 pages collected from {www.mayoclinic.org}. Each page also has regular sections, such as \textit{Symptoms}, \textit{Causes}, \textit{Risk Factors}, \textit{Treatments/Drugs}, and \textit{Prevention}, corresponding to \cd{has\-Symptom}, \cd{has\-Cause}, \cd{risk\-Factor}, \cd{has\-Treatment}, and \cd{pre\-ven\-tion\-Factor}, respectively.  
These corpora are all entity centric, i.e., each pages discusses a single entity.

\iflong{DailyMed includes information about treated diseases, adverse effects, drug ingredients, etc, and WikiDisease includes information about disease causes, treatments, symptoms, etc.}

We use GDep \cite{sagae:2007b}, a dependency parser trained on GENIA Treebank, to parse the corpora, followed by a simple POS-tag based chunker to extract NPs. We also extract a list (e.g. ``stomach upset, nausea, and dizziness'') for each coordinating conjunction that modifies a noun. For each NP mention, we extract features (described below) from its sentence; and for each coordinate list, we extract the similar features and the NP chunks included in it. A mention not inside a list is regarded as a singleton list that contains only one item.

\subsection{Mention Preparation}
\label{sec:example}
Relation mention sets, i.e. $R_s$ and $R_t$, are prepared with distant supervision.
The extracted NP mentions are distantly labeled using relation seed triples from Freebase (e.g. \cd{sideEffect(meloxican,nausea)}). Specifically, we require that the title entity matches the first argument value of the relation, and the NP mention matches the second argument value.
To improve the quality of $R_s$, we also require that the section from which the mention was taken is relevant to the relation. E.g., a mention labeled with the \cd{side\-Effect} relation must appear in a section entitled \textit{Side Effects}. Such constraint limits the number of mentions in $R_s$. In the next section, we will show how to extend this small but accurate example set to a larger training set of examples, with reasonable quality.

The concept mentions are designed to have high recall with respect to possible argument values for a relation.
For each relation $r$, we generate a set of concept mentions which lie in the range of $r$'s second argument. Following the DIEL system \cite{bing-EtAl:2015:EMNLP}, we extract concept instances from Freebase as seeds, and extend the seed set using LP in each corpus. The reached coordinate-term lists and singleton lists (NPs) are collected as concept mentions.
Thus, we get two concept mention sets: $C_s$ from the structured corpus, and $C_t$ from the target corpus. Note that some mentions in $C_s$ may come from unrelated sections; for instance, $C_s$ for the \cd{Symptom} concept
may contain mentions from the \textit{Overdose} section, which cannot be examples of the \cd{sideEffect} relation. Therefore, we filter out the mentions in $C_s$ that are not from the appropriate section for this concept.

We emphasize that the section-specific processing is \emph{only done on the structured corpus}, i.e. for $C_s$ and $R_s$.  Our target corpora have thousands of section titles, most of which are not related in any way to the relations being extracted. Thus the target relation mentions ($R_t$) and target concept mentions ($C_t$) are collected without considering section information.

\subsection{Relation Label Propagation}
\label{sec:lp_graphs}
With the relation mentions and the concept mentions lying in the range of the corresponding relation, we are able to distill a cleaner set of training relation examples to learn extractors.
$R_s$ contains more confident relation examples because of constraints by document structure, but it is limited in size. In contrast, the number of $R_t$ mentions is larger, but they are noisier. In general, the degree to which $R_t$ mentions will be useful may be domain- and corpus-specific.
$C_s$ and $C_t$ are generated with respect to the type of the mentions, but not their relationship with the title entity: e.g., a mention in $C_t$ corresponding to the NP ``dizziness'' would not be associated with the triple \cd{side\-Effect(meloxican,dizziness)}; and indeed, dizziness might be a condition treated by, not caused by, the title entity ``meloxican''. Therefore, $C_t$ itself cannot be directly used as relation examples, however, it can serve as a resource to distill relation examples.
In our experiments, $R_s$ mentions are always used as seed relation examples in LP, but we build bipartite propagation graphs with different combinations of the four sets of mentions and study their performance.

In total, we have 7 bipartite graphs, each with a different set of mentions from the following combinations: $R_s\cup C_s\cup R_t\cup C_t$, $R_s\cup C_s\cup R_t$, $R_s\cup C_s\cup C_t$, $R_s\cup C_s$, $R_s\cup R_t\cup C_t$, $R_s\cup R_t$, or $R_s\cup C_t$.
In a bipartite graph, one set of nodes are mentions, and the other set of nodes are features of mentions. An edge is added between each feature and each mention containing that feature. The edges are TFIDF-weighted (treating the features as words and the mentions as documents). Figure \ref{fig:bipartite} shows such a bipartite graph (edge weights are omitted), which has four mentions on the left-hand side, and eight features on the right-hand side.

\begin{figure}[t]
\centering
\includegraphics[width=2in]{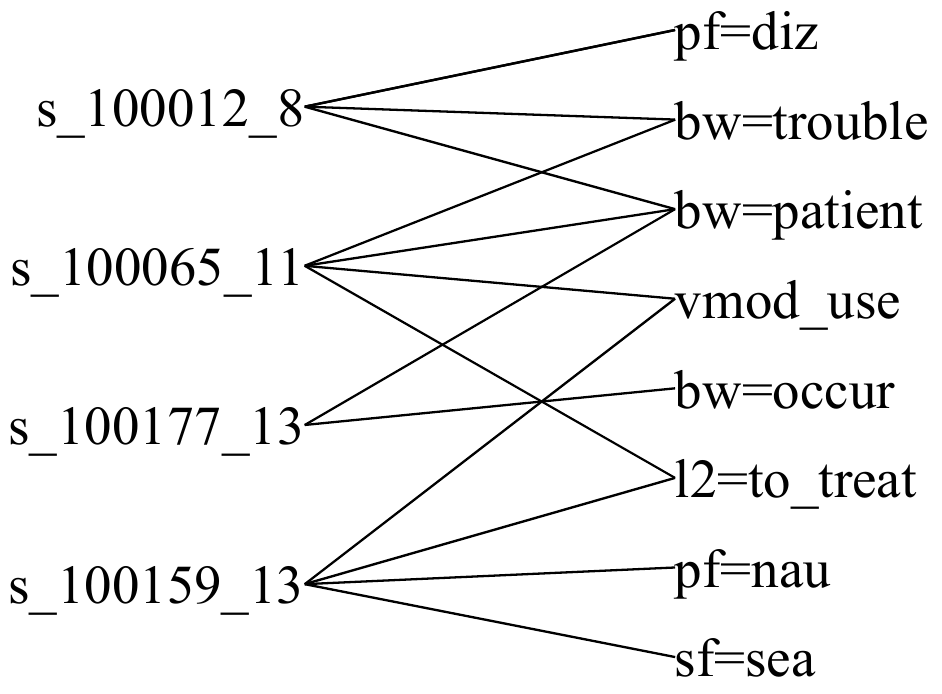}
%\vspace{-2mm}
\caption{\label{fig:bipartite}A bipartite graph of relation label propagation, with unique IDs on the left
and features on the right (refer to Section \ref{sec:learning}).}
\vspace{-4mm}
\end{figure}

We use an existing multi-class label propagation method,
namely, MultiRankWalk (MRW) \cite{DBLP:conf/asunam/LinC10}, which is a graph-based SSL method related to personalized
PageRank (PPR) \cite{Haveliwala03ananalytical} (aka random walk with restart  \cite{DBLP:conf/icdm/TongFP06}).
MRW can be viewed simply as computing a personalized PageRank vector for each class, each of which is computed using a personalization vector that is initially uniform over the seeds, and finally assigning to each node the class associated with its highest-scoring vector.  MRW's final scores depend on the centrality of nodes, as well as their proximity to seeds.
The MRW implementation we use is based on ProPPR \cite{wang2013programming}.

\subsection{Classifier Learning}
\label{sec:learning}
Given the ranked mention lists of these relation labels from the above LP, we pick the top $N$ to train binary classifiers, which can then be used to classify the entity mentions (singleton lists) and coordinate lists in a new document. We use the same feature generator for both mentions and lists.  Shallow features include: tokens in the NPs, and character prefixes/suffixes of these tokens; BOW from the sentence containing the NP; and tokens and bigrams from a window around the NPs.  From dependency parsing, we find the verb which is closest ancestor of the head of current NP, all modifiers of this verb, and the path to this verb.  For lists, the dependency features are computed relative to the head of the list.

We use SVMs \cite{CC01a} and discard singleton features, as well as the most frequent 5\% of features (as a stop-wording variant).
%In practical application, we will inevitably encounter a lot of ``other'' lists, i.e.,
%not belonging to the interested instance types in this task.
%However, we do not have other examples for training such a classifier.
Specifically, binary classifiers are trained with examples of one relation as the positives, and examples
of the other classes as negatives. We also add N general negative examples,
randomly picked from those that are not distantly labeled by any relation.
A linear kernel and default values for all other parameters are used \footnote{https://www.csie.ntu.edu.tw/~cjlin/libsvm/}.
A threshold 0.5 is used to cut positive and negative predictions.
If a new list or mention is not classified as positive by any classifier, it is predicted as ``other''.

\section{Experiments}

\subsection{Evaluation Dataset}
Our evaluation dataset contains 20 manually labeled pages, 10 pages each from the disease corpus WikiDisease and the drug corpus DailyMed. This data was originally generated in \cite{bing-aaai:2016}.  The annotated text fragments are manually chunked NPs which are the second argument values of any of the eight relations considered here, with the title
drug or disease entity of the corresponding document as the relation
subject. The evaluation data contains 436 triples for the disease domain
and 320 triples for the drug domain.
A system's task then is to extract all correct values of the second argument of a given relation from a test document.

\iflong{This is a test of the system's performance in extracting information from particular documents.

Our second evaluation dataset is composed of questions in the training dataset of BioASQ3-Task
B \cite{bioasq}.
We consider only the factoid and list questions, which require a particular entity
name (e.g., a disease) or a list of entities as an answer.  From BioASQ3 we extracted 58 questions related to our 8 relations, including 37 factoid ones and 21 list ones.
Each natural language question is translated to a structured database query,
which can be tested on any KB. For instance, query(treatsDisease, daonil, Y)
expects a disease name as answer, e.g. ``diabetes mellitus''.  This is a test of the overall coverage of the extracted KB.}

\subsection{Experimental Comparisons}
The first three baselines are distant supervision (DS) systems.
They classify each testing NP mention into one of the interested
relation types or ``other'', using naive matching to the Freebase seed triples as distant
supervision. Each sentence in the corpus is processed with the
same preprocessing pipeline to detect NPs. Then,
these NPs are labeled with the Freebase seed triples.
The features are defined and extracted in the
same way as we did for DIEJOB, and binary
classifiers are trained with the same method.
The first DS baseline, named \textit{DS\_Struct}, only uses the section-filtered
examples from a structured corpus, i.e. $R_s$, as training data.
The second DS baseline, named \textit{DS\_Target}, only uses labeled
examples from the target corpus, i.e. $R_t$.
While the third DS baseline, named \textit{DS\_Both}, uses examples from both target corpus and structured
corpus.

We also compare against two latent variable learners. The first  is \textit{MultiR} \cite{hoffmann2011knowledge} which models each relation mention separately and aggregates their labels using a deterministic OR. The second one is \textit{MIML-RE} \cite{Surdeanu:2012:MML:2390948.2391003} which has a similar structure to \textit{MultiR}, but uses a classifier to aggregate the mention level predictions into an entity pair prediction. We used the publicly available code from the authors\footnote{http://aiweb.cs.washington.edu/ai/raphaelh/mr/ and http://nlp.stanford.edu/software/mimlre.shtml} for our experiments. Since these methods do not distinguish between structured and unstructured corpora, we used the union of these corpora in our experiments, and the feature set used in the bipartite graph. We found that the performance of these methods varies significantly with the number of negative examples used during training, and hence we tuned these and other parameters\footnote{Parameters include the number of epochs (for both \textit{MultiR} and \textit{MIML-RE}) and the number of training folds for \textit{MIML-RE}.} directly on the evaluation data, and report their best performance. Another distant-supervision baseline we compare to is the \textit{Mintz++} model from \cite{Surdeanu:2012:MML:2390948.2391003}, which improves on the original model from \cite{mintz2009distant} by training multiple classifiers, and allowing multiple labels per entity pair.

\iffalse
\begin{figure}
\centerline{\includegraphics[width=\linewidth]{./fig/graph_section_link}}
%\vspace{-0.1in}
\caption{An example of label propagation graph in \cite{bing-aaai:2016}. }
\label{fig:graph_section_link}
\end{figure}
\fi

We also compare with DIEBOLDS \cite{bing-aaai:2016}, which uses LP on a graph containing entity mention pairs.  The graph used by DIEBOLDS is more complex than the mention-feature graph used here, in DIEJOB.  One set of vertices correspond to (title-entity, mention-entity) pairs. The other set of vertices are identifiers for coordinate lists: a mention pair is connected with the lists from any document describing the subject, and containing the mention. Additional edges are also introduced based on document structure and BOW context features.\iflong{\footnote{DIEBOLDS adds an edge for two subject-NP pairs if their NP strings match and the two NPs come from the same section of two documents. If the NPs of pairs appear in the same section of a document, these pairs are connected. DIEBOLDS also adds a weighted edge for two pairs if their NPs have similar features, as calculated with TFIDF weighted BOW of contexts for the NPs in all sentences they appear.}}  DIEBOLDS performs label propagation from the mention pairs distantly labeled with Freebase relation triples.

\subsection{Experimental Settings}
%\bc{these numbers are matched triples of the input corpus, after seconsString, so recheck the number of triples from Freebase, use them to replace numbers here}

We extracted triples of the eight relations from Freebase as distant labeling seeds. Specifically, if the subject of a triple matches with a drug or disease name
in a corpus and its object value also appears in that document, it is extracted.
For the disease domain, we get 2022, 2453, 905, 753, and 164
triples for \cd{has\-Treatment}, \cd{has\-Symptom}, \cd{risk\-Factor}, \cd{has\-Cause}, and
\cd{pre\-ven\-tion\-Factor}, respectively. For the drug domain, we get 3112, 315, and 265
triples for \cd{used\-To\-Treat}, \cd{con\-di\-tions\-This\-May\-Pre\-vent}, and
\cd{side\-Effect}, respectively.

We have two strategies to pick the top $N$ lists for classifier learning. One strategy picks the top $N$ directly, without distinguishing if they come from the structured corpus or the target corpus. It is referred to as \textit{DIEJOB\_Both}. The other strategy picks the top N examples only from the target corpus, and it is referred to as \textit{DIEJOB\_Target}. Here our concern is the difference between the feature distributions of the two corpora.

\iffalse
DIEBOLDS and our DIEJOB can classify both singleton and coordinate lists.
After that, lists are broken into items, i.e. NPs, for evaluation. \fi
We evaluate the performance of different systems
from an IR perspective: a title entity (i.e., document name)
and a relation together act as a query, and the extracted NPs as retrieval results.

\subsection{Results on Labeled Pages}
The results for precision, recall and F1 measure
are given in Table \ref{t:prf}.
The results for DIEBOLDS are from \cite{bing-aaai:2016}.
The systems with ``*'' are directly tuned on the evaluation data and should be considered as upper bounds on true performance.
DIEJOB\_Target and DIEJOB\_Both are tuned with a tuning dataset (details in Section~\ref{sec:variants}).
(Note that for the disease domain, DIEJOB\_Both and DIEJOB\_Both* get the same results, because
they use the same parameters, although they are tuned with different data.)

\begin{table}[!t]
\center{
\caption{Extraction results on the labeled pages.\label{t:prf}}
\begin{small}
\begin{tabular}{@{}c@{~}|@{~}c@{~}@{~}c@{~}@{~}c@{~}|@{~}c@{~}@{~}c@{~}@{~}c@{}}
  \hline
 &  & Disease &  &  & Drug & \\
\cline{2-7}
& P &    R   & F1 & P & R & F1 \\
\hline
  \hline
DS\_Struct & 0.300 & 0.300 & 0.300 & 0.232 & 0.072 & 0.110 \\
DS\_Target & 0.228 & 0.335 & 0.271 & 0.170 & 0.188 & 0.178 \\
DS\_Both & 0.233 & 0.353 & 0.281 & 0.154 & 0.175 & 0.164 \\
\hline
%Freebase & 0.202 &  0.037 &  0.062 & 0.318 & 0.022 & 0.041 \\
DIEBOLDS & 0.143 & 0.372 & 0.209 & 0.050 & 0.435 & 0.090  \\
MultiR* & 0.198 & 0.333 & 0.249 & 0.156 & 0.138 & 0.146  \\
Mintz++* & 0.192 & 0.353 & 0.249 & 0.177 & 0.178 & 0.178  \\
MIML-RE* & 0.211 & 0.360 & 0.266 & 0.167 & 0.160 & 0.163  \\
\hline
DIEJOB\_Target & 0.231 & 0.337 & 0.275 & 0.299 & 0.300 & 0.300 \\
DIEJOB\_Both & 0.317 & 0.333 & \textbf{0.324} & 0.327 & 0.288 & \textbf{0.306} \\
\hline
DIEJOB\_Target* & 0.235 & 0.339 & 0.277 & 0.289 & 0.425 & 0.344 \\
DIEJOB\_Both* & 0.317 & 0.333 & 0.324 & 0.282 & 0.422 & 0.338 \\
\hline
\end{tabular}
\end{small}
}
%\vspace{-0.2cm}
\vspace{-5mm}
\end{table}

\begin{figure*}[!t]
\begin{subfigure}{.5\textwidth}
  \centering
  \includegraphics[width=\linewidth]{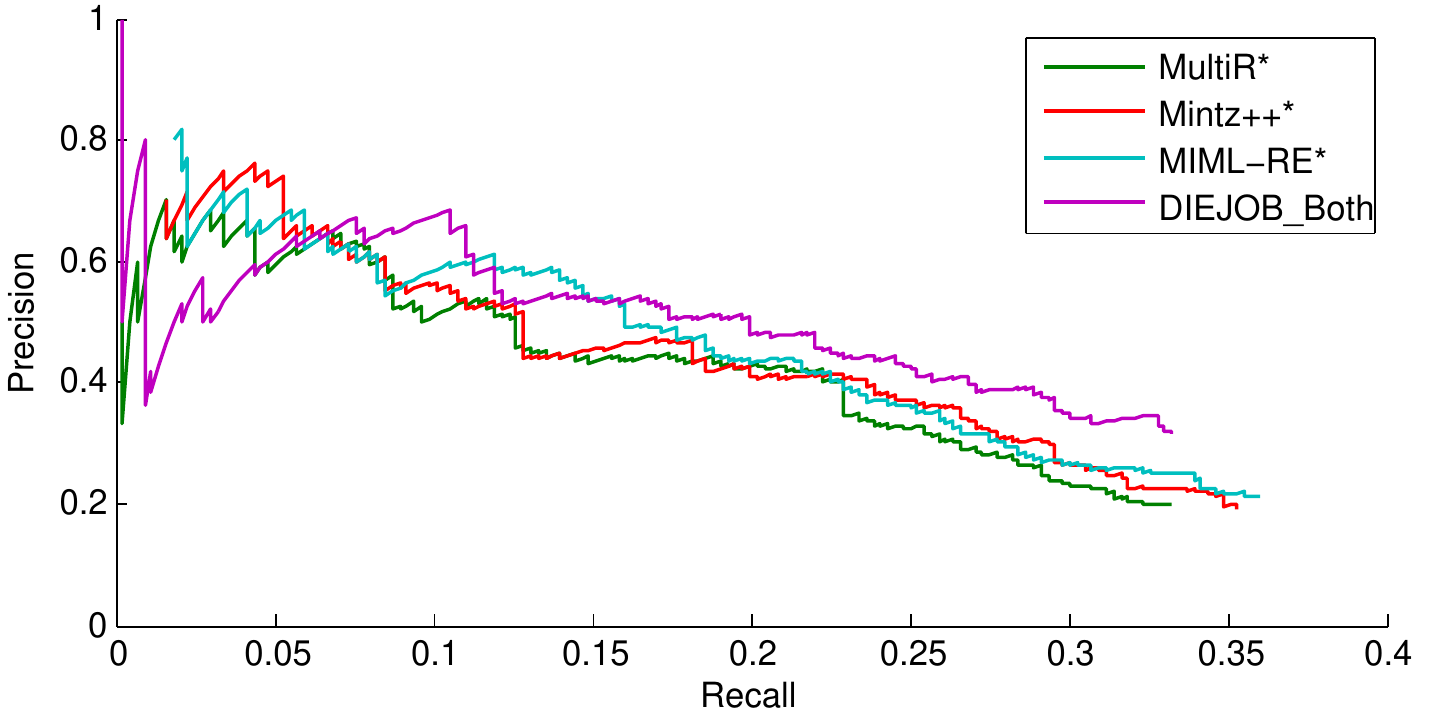}
  \caption{Disease domain.}
  \label{fig:disease_pr}
\end{subfigure}%
\begin{subfigure}{.5\textwidth}
  \centering
  \includegraphics[width=\linewidth]{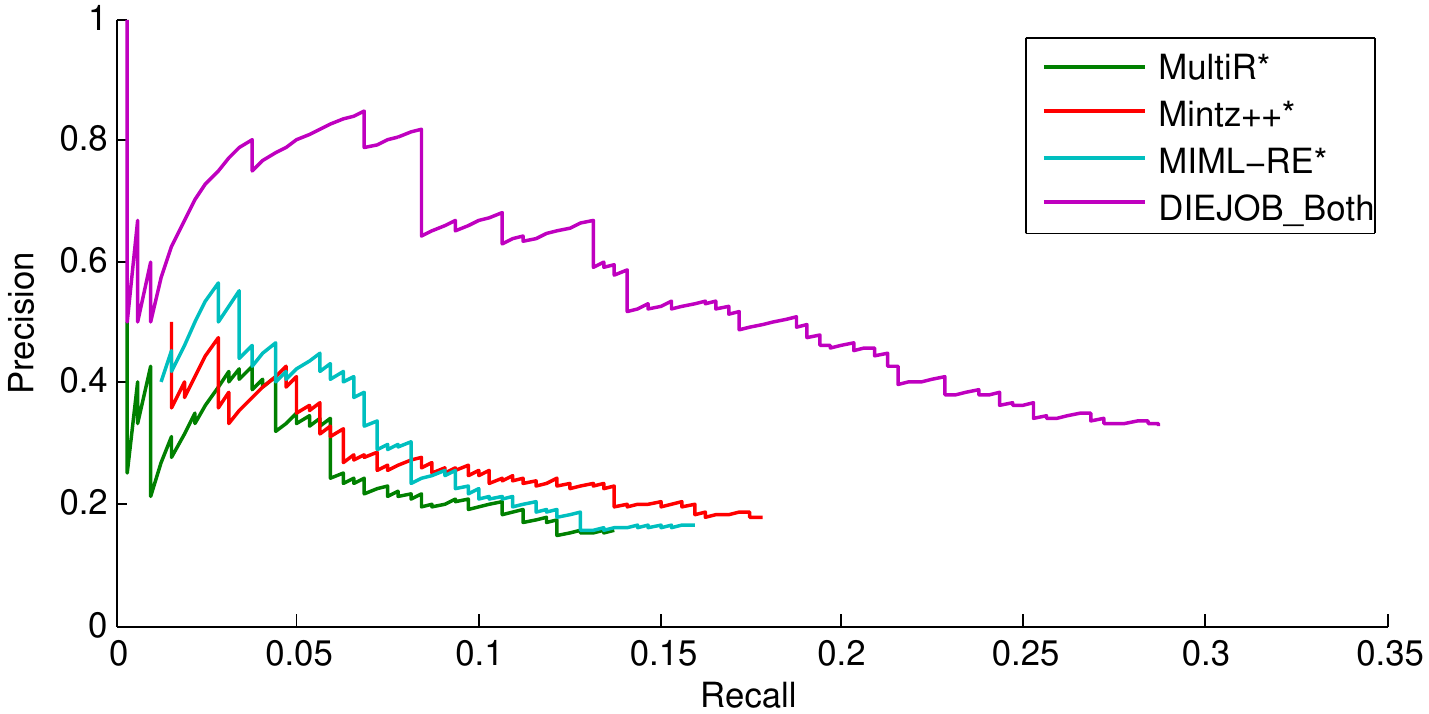}
  \caption{Drug domain.}
  \label{fig:drug_pr}
\end{subfigure}
\vspace{-2mm}
\caption{\label{fig:pr_curve}Precision-recall curves.}
\vspace{-5mm}
\end{figure*}

DIEJOB\_Both outperforms all the other systems. Compared with MultiR,
Mintz++, and MIML-RE, the relative improvements under the F1 measure are 22\% to 30\% in the disease domain,
and 72\% to 110\% in the drug domain. The precision values of DIEJOB\_Both are
much higher than previous work.  For recall, DIEBOLDS and DIEJOB\_Both's performance are comparable to the latent-variable systems on the disease domain and much better on the drug domain. One reason may be that our method
predicts one label for a coordinate-term list (lists are common in the drug domain), which implicitly coordinates the
labels of list items, while MultiR, Mintz++, and MIML-RE break a list
into individual items which are predicted separately.

The precision values of DIEBOLDS are much lower than DIEJOB, especially for the drug domain. Unlike DIEJOB, DIEBOLDS builds an LP graph containing all singleton and coordinate lists of noun phrases in the corpus, which introduces many irrelevant examples.  DIEBOLDS achieves the highest recall values, but in practice, it is also likely to predict a testing mention as belonging to one of the eight relations, but not ``other''.

On these tasks, the simple DS baselines' performance is competitive with MIML-RE and the
other complex models. One exception is DS\_Struct
on the drug domain, where the recall is only 0.072.
This is perhaps because the total number of examples in $R_s$ for the three drug relations is only 485, which is too small
to get good recall. Interestingly, the precision
of DS\_Struct is better than DS\_Target and DS\_Both for both domains,
presumably, because of the high quality of the examples in $R_s$.

For the disease domain, DIEJOB\_Both performs better than
DIEJOB\_Target, no matter how they are tuned (i.e. on tuning or evaluation data).
This shows that the mentions from $R_s$ and $C_s$ of MayoClinic corpus provide good training examples.
For the drug domain, DIEJOB\_Both and DIEJOB\_Target achieve similar
results. This may be because DIEJOB\_Both is more sensitive to the difference
in feature distributions of structured and target corpora,
since it uses examples from the structured corpus to learn
classifiers as well. Among the four corpora we use, WebMD, MayoClinic,
and WikiDisease are written to be readable by a large
audience, while DailyMed articles are more difficult in terms of
readability: hence the difference between the structured and
unstructured corpora is larger in the drug domain.

\iflong{
We also employ Freebase as a comparison system, using the relevant facts from Freebase as the system output.
It is not unexpected to observe very low recall values in both domains, since the coverage
of Freebase on specific domains such as biomedical domain is limited.
However, the precision values are also surprisingly low, especially for disease domain.
The reason is two-fold. First, our labeled pages do not contain all true facts relevant to a title entity, and true facts contained by Freebase but not not mentioned on the pages are considered as false positives in our evaluation. The second reason is that a number of the facts in Freebase (in these subdomains) actually appear to be incorrect.}

Precision-recall curves are given in Figure \ref{fig:pr_curve}.
For the drug domain, DIEJOB's precision is consistently better,
at the same recall level, than any of the other methods. For the disease domain,
our system's precision is generally better after the recall level 0.05.

\subsection{Tuning and Variant Comparison}
\label{sec:variants}
Here we examine the performance of different variants, and the effect of the parameter $N$.
The performance of all graph variants on a tuning dataset (containing 10 labeled pages) is given in Figure \ref{fig:all_variants}. Combined with the strategies for picking top N (i.e. DIEJOB\_Target and DIEJOB\_Both), there are 13 variants: shown in Figures \ref{fig:diejob-b_disease_variants} and \ref{fig:diejob-t_disease_variants} for disease; Figures \ref{fig:diejob-b_drug_variants} and \ref{fig:diejob-t_drug_variants} for drug. Note that DIEJOB\_Target does not have the variant $R_sC_s$, because $R_sC_s$ does not contain any examples from the target corpus.

For the disease domain, the same variant under DIEJOB\_Both and DIEJOB\_Target performs similarly, and on average, DIEJOB\_Both is slightly better than DIEJOB\_Target. For the drug domain, on average, DIEJOB\_Target is better than DIEJOB\_Both. One explanation is that the two corpora in disease domain are similar in the aspect of feature distribution, so in general, mixing the examples from them are beneficial. However, the effect of such a mixture is negative for drug domain, whose structured and target corpora are dissimilar.

In Table \ref{t:prf}, the reported results of the tuned DIEJOB\_Both and DIEJOB\_Target for the disease domain are from the variants $R_sC_s$ and $R_sC_sR_t$ respectively, while for drug domain, both are from $R_sR_t$.
One explanation could be: (1) if the structured corpus is similar to the target corpus, it is better to use DIEJOB\_Both, and including examples of the structured corpus (e.g., $R_sC_s$ and $R_sC_sR_t$, both have $C_s$ used) generally performs well with a larger N value; (2) if the structured and target corpora are dissimilar, DIEJOB\_Target is better and $R_sR_t$ has an advantage over other variants where the main focus is distilling good training examples from $R_t$ and a smaller number of top N examples is preferred.

\begin{figure*}[!t]
\begin{subfigure}{\textwidth}
  \centering
  \includegraphics[width=0.6\linewidth]{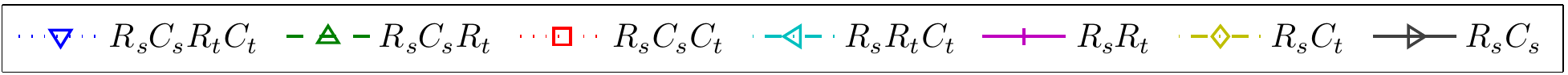}
  \label{fig:sfig1}
\end{subfigure}% 
\newline
\begin{subfigure}{.5\textwidth}
  \centering
  \includegraphics[width=0.85\linewidth]{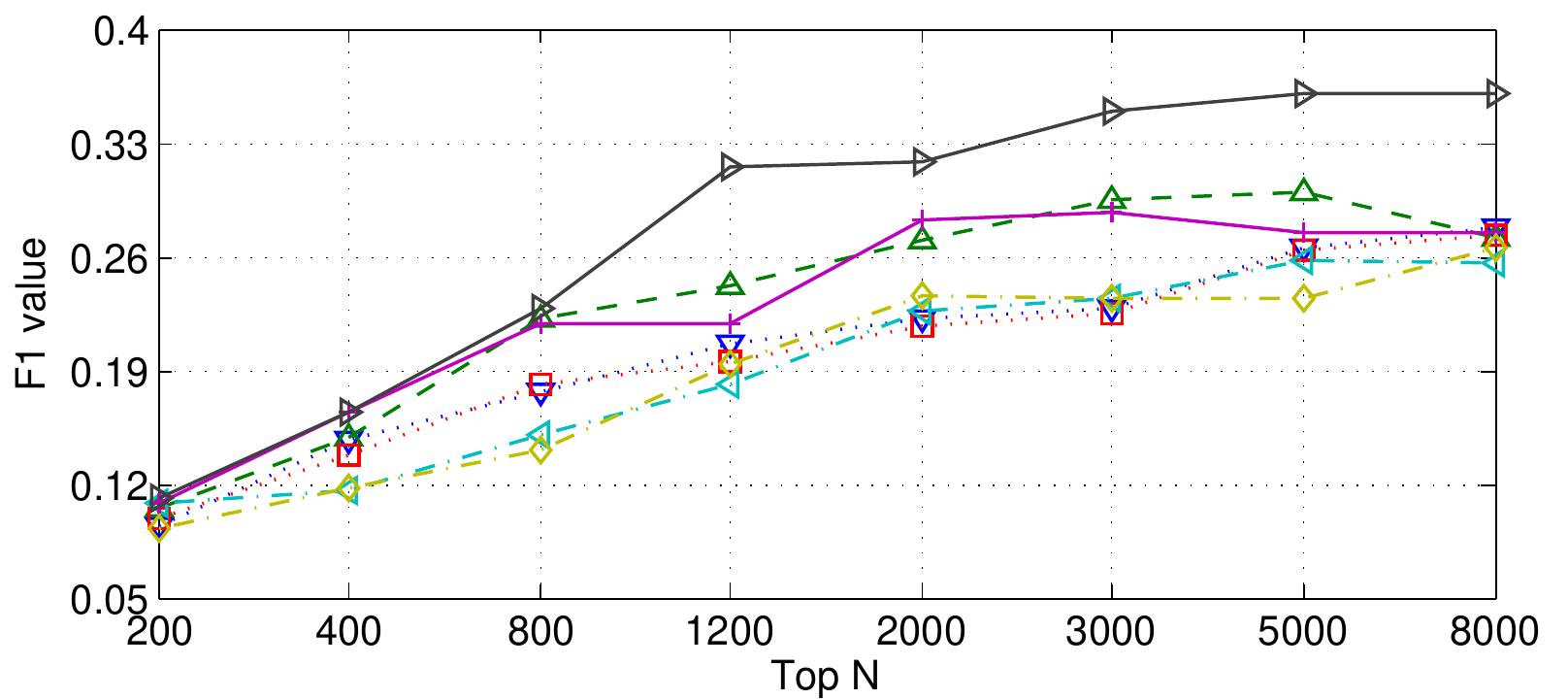}
  \caption{\label{fig:diejob-b_disease_variants}Variants of DIEJOB\_Both for disease domain.}
  \label{fig:sfig2}
\end{subfigure}
\begin{subfigure}{.5\textwidth}
  \centering
  \includegraphics[width=0.85\linewidth]{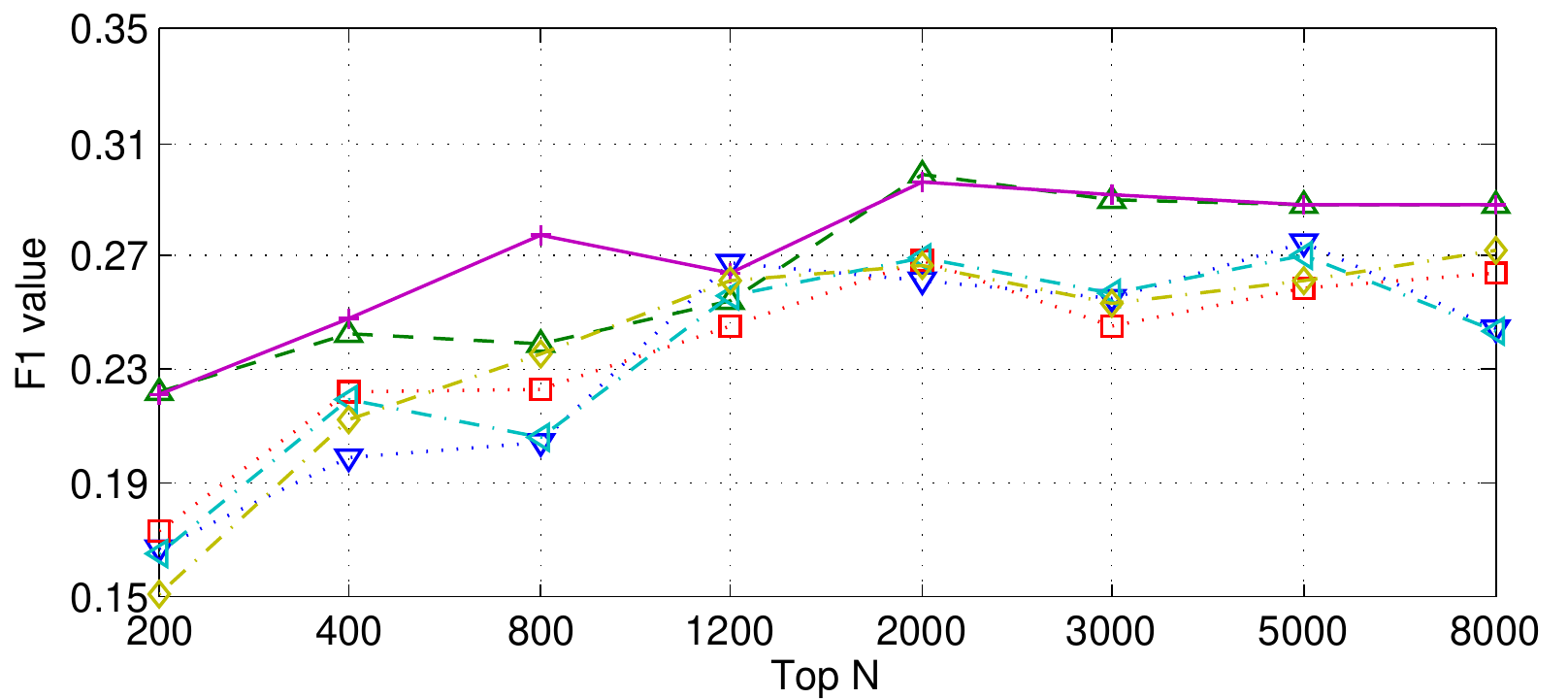}
  \caption{\label{fig:diejob-t_disease_variants}Variants of DIEJOB\_Target for disease domain.}
  \label{fig:sfig3}
\end{subfigure}
\begin{subfigure}{.5\textwidth}
  \centering
  \includegraphics[width=0.85\linewidth]{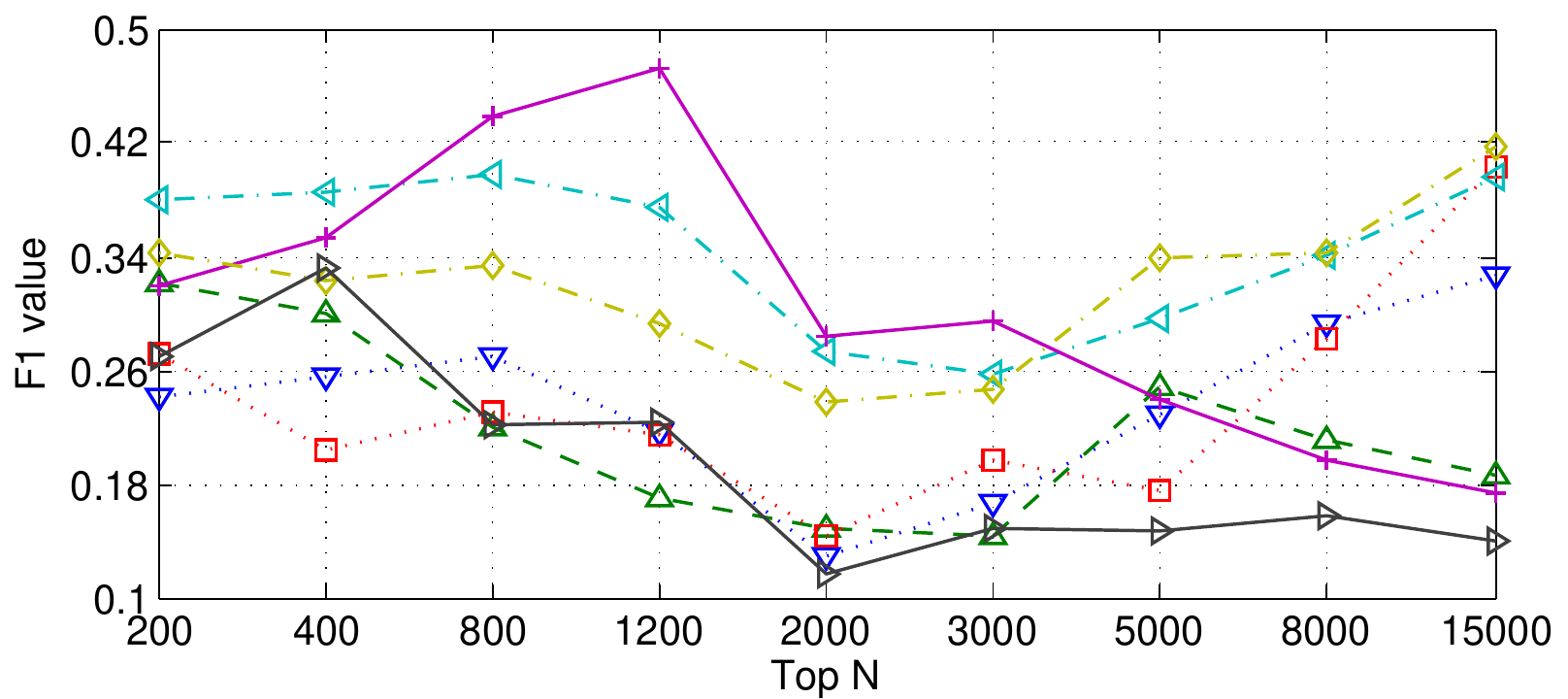}
  \caption{\label{fig:diejob-b_drug_variants}Variants of DIEJOB\_Both for drug domain.}
  \label{fig:sfig4}
\end{subfigure}%
\begin{subfigure}{.5\textwidth}
  \centering
  \includegraphics[width=0.85\linewidth]{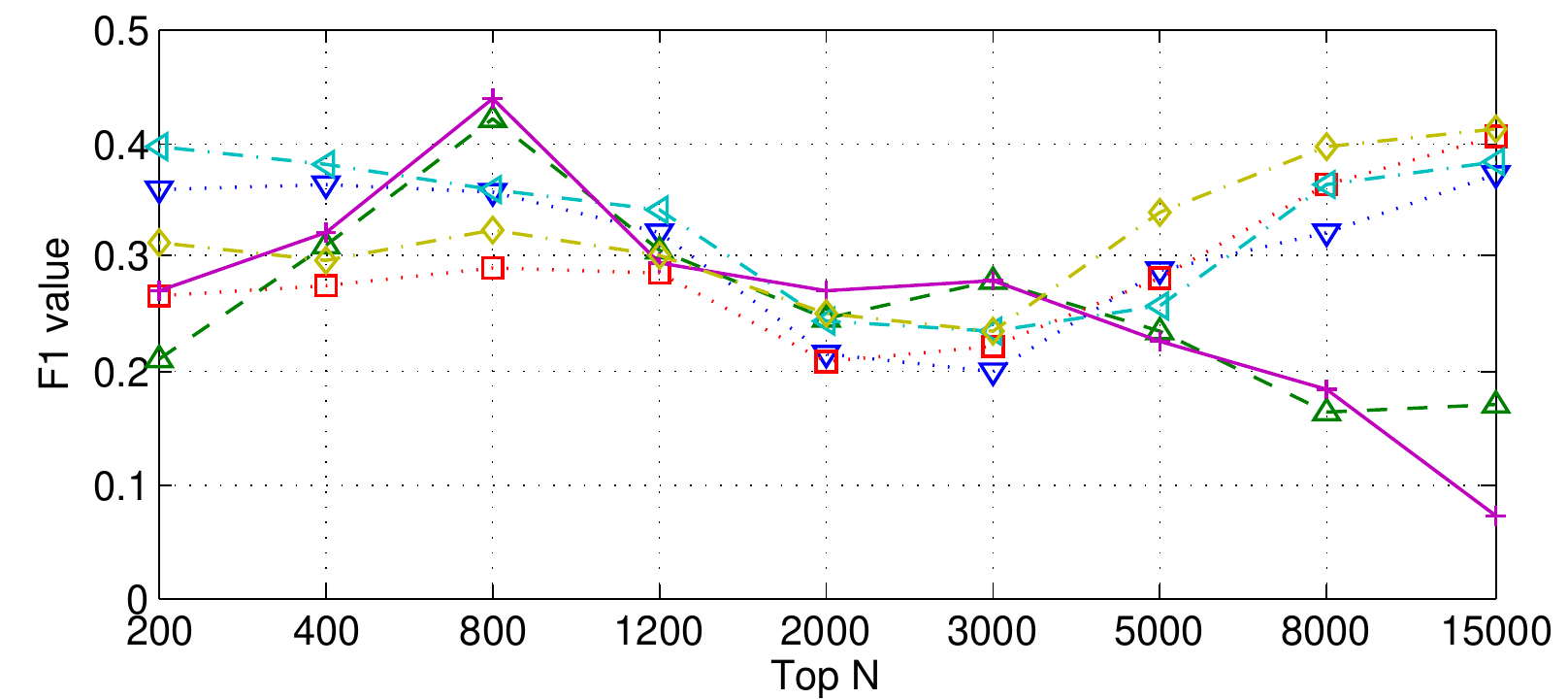}
  \caption{\label{fig:diejob-t_drug_variants}Variants of DIEJOB\_Target for drug domain.}
  \label{fig:sfig5}
\end{subfigure}
\caption{\label{fig:all_variants}Performance of DIEJOB variants and effect of the parameter N.}
\vspace{-5mm}
\end{figure*}

\section{Related Work}

\vspace{-2mm}

\iflong{Distant supervision was initially employed by \cite{CravenISMB99}
in the biomedical domain under a
different terminology, i.e. \emph{weakly labeled data}.
Then, it attracted attentions from the IE community.
%\cite{wu2007autonomously,wu2010open,mintz2009distant,riedel2010modeling,hoffmann2011knowledge,Surdeanu:2012:MML:2390948.2391003}.
\cite{mintz2009distant} employed Freebase to label sentences containing
a pair of entities that participate in a known Freebase relation, and
aggregated the features from different sentences for this relation.
\cite{wu2007autonomously,wu2010open}
also proposed an entity-centric corpus oriented method, and
employed infoboxes to label their corresponding Wikipedia articles.
DIEJOB employs Freebase to label entity-centric documents of particular domains.}

To overcome the noise in distantly-labeled examples, \cite{riedel2010modeling} introduced an ``at least one'' heuristic, where instead of taking all mentions for a pair as correct examples only at least one of them is assumed to express that relation.
MultiR \cite{hoffmann2011knowledge} and MIML-RE
\cite{Surdeanu:2012:MML:2390948.2391003} extend this approach
to support multiple relations expressed by different
sentences in a bag. Unlike these approaches, DIEJOB improves the quality of training data with a bootstrapping step before feeding the noisy examples into a learner, by using the confident examples from a structured corpus as seeds. The benefit of this step is two-fold. First,
it distills the distantly-labeled examples by propagating labels from good seed examples,
and downweights the noisy ones. Second, the propagation will
walk to more relation examples in the concept mention set that cannot be distantly labeled with triples from knowledge bases.

Document structure was previously explored by \cite{bing-aaai:2016}, which used the structure
to enrich an LP graph by adding coupling edges between mentions in the same section of particular documents.
In this work, we explore the semantic association between section titles and
relation arguments. Furthermore, we perform a joint bootstrapping on relation and type mentions
to collect training examples with better quality.
Technically, the propagation graphs used are different: DIEJOB's graph has carefully produced mention nodes (from those four sets) and their feature nodes, while DIEBOLDS’s graph has triple nodes (i.e., subject-NP pairs) and all singleton and coordinate lists of noun
phrases of the corpora.
Accordingly, their propagation seeds are different: DIEJOB uses confident examples as seeds (labeled from particular sections of a structured corpus) to propagate labels to more examples via feature similarity, while DIEBOLDS directly uses Freebase triples as seeds and propagates labels through edges built from coordinate lists and sections.

In the classic bootstrap learning scheme \cite{riloff99Learning,agichtein00snowball,DBLP:conf/acl/BunescuM07},
a small number of seed instances are used to
extract new patterns from a large corpus, which are then used to extract more
instances. Then in an iterative fashion, new instances are used to extract more patterns.
DIEJOB departs from earlier bootstrapping methods in combining label
propagation with a standard classification learner,
and it can improve the quality of distant examples and collect new examples
simultaneously.

\section{Conclusions}
\vspace{-3mm}
We proposed the DIEJOB framework to generate good examples for distantly-supervised IE. It exploits the document structure of a small well-structured corpus to collect seed relation examples, and it also collects concept mentions that could be the second argument values of relations. DIEJOB then conducts label propagation to find mentions that can be confidently used as training examples to train classifiers for labeling new entity mentions. The experimental results show that this approach consistently and significantly outperforms state-of-the-art approaches.

\iflong{
For future work, one direction is to investigate the relationship among the extracted relations. Another direction is to explore better features, e.g. embeddings, for such domain-specific IE task. As shown by the results, a small structured corpus is already helpful for the performance, so another promising direction is to supplement the system to also process high-quality and large-quantity data available in semi-structured documents such as web tables.}

%\section{Acknowledgments}
%This work was funded by a grant from Baidu USA and by the NSF under research grant IIS-1250956.
\section{Acknowledgments}
This work was funded by grants from Baidu USA and Google, and the research grant IIS-1250956 from NSF.

\bibliographystyle{acl}
\balance
\bibliography{all}

\end{document}